\documentclass[conference]{IEEEtran}
\IEEEoverridecommandlockouts
\usepackage{cite}
\usepackage{amsmath,amssymb,amsfonts}
\usepackage{graphicx}
\usepackage{textcomp}
\usepackage{xcolor}
\usepackage{bm}
\usepackage{tabularx}
\usepackage{multirow}
\usepackage{tikz}
\usetikzlibrary{positioning, calc, fit, shapes, arrows, external}
\usepackage{pgfplots}
\pgfplotsset{compat=1.17}
\usepgfplotslibrary{fillbetween}

\usepackage{mathtools}
\usepackage[caption=false]{subfig}
\usepackage [english]{babel}
\usepackage [autostyle, english = american]{csquotes}
\MakeOuterQuote{"}

\usepackage{algorithm}
\usepackage{algpseudocode}

\def\BibTeX{{\rm B\kern-.05em{\sc i\kern-.025em b}\kern-.08em
    T\kern-.1667em\lower.7ex\hbox{E}\kern-.125emX}}
\begin{document}

\title{Reliable Probability Intervals For Classification Using Inductive Venn Predictors Based on Distance Learning}
\author{}

% \author{\IEEEauthorblockN{Dimitrios Boursinos}
% \IEEEauthorblockA{Institute for Software Integrated Systems \\
% Vanderbilt University\\
% Nashville TN, USA \\
% dimitrios.boursinos@vanderbilt.edu}
% \and
% \IEEEauthorblockN{Xenofon Koutsoukos}
% \IEEEauthorblockA{Institute for Software Integrated Systems \\
% Vanderbilt University\\
% Nashville TN, USA \\
% xenofon.koutsoukos@vanderbilt.edu}
% }

\author{
\IEEEauthorblockN{Dimitrios Boursinos and Xenofon Koutsoukos}
\IEEEauthorblockA{Institute for Software Integrated Systems \\
Vanderbilt University\\
Nashville TN, USA \\
    \{dimitrios.boursinos, xenofon.koutsoukos\}@vanderbilt.edu}\\%
}
% \IEEEpubid{978-1-6654-3156-9/21/\$31.00 ©2021 IEEE}
 \IEEEpubid{\makebox[\columnwidth]{978-1-6654-3156-9/21/\$31.00~\copyright2021 IEEE \hfill} \hspace{\columnsep}\makebox[\columnwidth]{ }}

\maketitle

\IEEEpubidadjcol

\begin{abstract}
Deep neural networks are frequently used by autonomous systems for their ability to learn complex, non-linear data patterns and make accurate predictions in dynamic environments. However, their use as black boxes introduces risks as the confidence in each prediction is unknown. Different frameworks have been proposed to compute accurate confidence measures along with the predictions but at the same time introduce a number of limitations like execution time overhead or inability to be used with high-dimensional data. In this paper, we use the Inductive Venn Predictors framework for computing probability intervals regarding the correctness of each prediction in real-time. We propose taxonomies based on distance metric learning to compute informative probability intervals in applications involving high-dimensional inputs. Empirical evaluation on image classification and botnet attacks detection in Internet-of-Things (IoT) applications demonstrates improved accuracy and calibration. The proposed method is computationally efficient, and therefore, can be used in real-time.
\end{abstract}
\begin{IEEEkeywords}
deep neural networks, assurance monitoring, inductive venn predictors, probability intervals
\end{IEEEkeywords}
\section{Introduction}
\label{sec:introduction}
% Motivation
Modern Deep Neural Network (DNN) architectures have the capacity to be trained using high-dimensional data and make accurate decisions in dynamic and uncertain environments. This ability makes them a common choice for many autonomous system applications. 
However, when DNNs are used as black boxes in  safety-critical systems, they may result in disastrous consequences if it is not possible to reason about their predictions.

% Recent advances in Deep Neural Network (DNN) architectures~\cite{krizhevsky2012imagenet,szegedy2015going,simonyan2014very} has made them a common choice for systems with various degrees of autonomy. Designs incorporating convolutional layers have increased the depth and width of the networks while allowing for real-time operation using Graphics Processing Units (GPUs). This allows them to make accurate predictions on high-dimensional data and handle the uncertainty and variability of the real-world. However, their integration as black boxes in trustworthy and safety-critical systems can have disastrous consequences when they cannot reason about their predictions.

% Hazards
The training of a Learning Enabled Component (LEC) requires specification of the task, performance measure for evaluating how well the task is performed, and experience in the form of training and testing data. An LEC, such as a DNN, during system operation exhibits some nonzero error rate and the true error rate is unknown and can only be approximated during design time using the available data. Confidence values, such as the softmax probabilities which are used by most DNNs for classification, are usually greater than the actual posterior probability that the prediction is correct. Important factors that make modern DNNs overconfident are the depth, width, and techniques like weight decay, and batch normalization~\cite{Guo:2017:CMN:3305381.3305518}.

% Problem considered in this paper
Our objective is to complement the predictions made by DNNs with a computation of confidence. The confidence can be expressed as probability intervals to characterize the correctness of the DNN prediction. An efficient and robust approach must ensure that the actual accuracy of a DNN is contained in the computed intervals and the width of the intervals is small. We focus on computationally efficient algorithms that can be used in real-time. The proposed approach is based on the Inductive Venn Predictors (IVP) framework~\cite{Vovk:2005:ALR:1062391}. IVP computes the probability intervals for an unknown input leveraging knowledge it has acquired from previous predictions on labeled data. Most of the IVP or Venn Predictors (VP) applications in the literature are evaluated on low-dimensional data~\cite{vovk2003self,Vovk:2005:ALR:1062391,lambrou2015inductive,papadopoulos2011reliable,lambrou2012reliable,papadopoulos2013reliable}. 

% The importance of reliable confidence metrics in safety-critical systems has been recognized by many researchers who proposed methods for LEC calibration based on different principles. 
The estimation of reliable predictive uncertainty has become an important part of many modern machine learning components used in safety-critical applications.
Even though many of the proposed methods produce well-calibrated models, their application in the real world is challenging. In ~\cite{pereyra2017regularizing,pmlr-v80-kumar18a}, new training algorithms and loss functions are proposed to achieve well-calibrated DNNs. 
% XK: Try not to be so negative for other approaches
These approaches require training DNN models from scratch and cannot be used with pre-trained ones. Another category of calibration methods like the Platt's scaling~\cite{Platt99probabilisticoutputs} and temperature scaling~\cite{Guo:2017:CMN:3305381.3305518} proposes ways of post-processing the outputs of already trained models to produce calibrated confidence measures. In~\cite{kumar2019verified,ovadia2019can}, it is shown that these methods are not as well-calibrated as it is reported especially when the validation data are not independent and identically distributed (IID) and in the presence of distribution shifts. The Conformal Prediction (CP) framework is developed to compute prediction sets to satisfy a desired significance level~\cite{Vovk:2005:ALR:1062391,Shafer:2008:TCP:1390681.1390693,balasubramanian2014conformal}. The confidence value assigned to each possible class is in the form of $p$-values which is less intuitive than estimating the confidence as probabilities. Another way of obtaining confidence information about predictions is by using algorithms based on the Bayesian framework. The use of this framework, however, require some prior knowledge about the distribution generating the data. In the real world, this distribution is unknown and it has to be chosen arbitrarily. In~\cite{10.5555/2016945.2016967}, it is shown that the predictive regions produced by Gaussian Processes, a popular Bayesian machine learning approach, may be incorrect and misleading when the correct prior is not known.

% only applied on binary classifiers~\cite{Platt99probabilisticoutputs}, computing confidence in terms of less intuitive p-values~\cite{papadopoulos2007conformal} and lack of evaluation on high-dimensional data~\cite{papadopoulos2013reliable,lambrou2015inductive}.
The main contribution of our work is that we compute low-dimensional, appropriate, embedding representations of the original inputs in a space where the Euclidean distance is a measure of similarity between the original inputs, in order to handle high-dimensional inputs in real-time. Then, we implement four different taxonomies that split the low-dimensional data into categories based on their similarity. Last, we present an empirical evaluation of the approach using two datasets for image classification problems with a large number of classes as well as detection of botnet attacks in an IoT device. The underlying models are chosen according to the input size and shape keeping into account the low-latency and low-power properties to meet the resource constraints of the variety of use cases~\cite{howard2017mobilenets}.

% In this paper, we investigate the IVP framework for assurance monitoring of systems containing machine learning components. The approach leverages IVP for providing predictions with well-calibrated confidence. The main contribution is that in order to handle high-dimensional inputs in real-time, we compute low-dimensional, appropriate, embedding representations of the original inputs. We implement various taxonomies that split the low-dimensional data into categories based on their similarity. A second contribution of the paper is that it presents an empirical evaluation of the approach using two datasets for image classification problems with large number of classes. The underlying model we chose to use is the MobileNet, a popular network architecture that provides low-latency and low-power models to meet the resource constraints of a variety of use cases~\cite{howard2017mobilenets}.
% \input{related_work}
\section{Problem}
\label{sec:problem}
A perception component in an autonomous system aims to observe and interpret the environment in order to provide information for decision-making. For example, a DNN can be used for classifying traffic signs in autonomous vehicles. The problem is to complement the prediction of the DNN with a computation of confidence. An efficient and robust approach must ensure a small and well-calibrated error rate to enable real-time operation. The approach must ensure a bounded small error rate while limiting the number of inputs for which an accurate prediction cannot be made. 

During system operation, for each new input a prediction is made, usually by a LEC and the objective is to compute a valid measure of the prediction's confidence. The objective is twofold: (1) provide guarantees for the error rate of the prediction and (2) limit the number of input examples for which a confident prediction 
cannot be made. Well-calibrated confidence in terms of probabilities can be used for decision-making, for example, by generating warnings when human intervention is required.

The Venn Prediction (VP) framework can produce predictions with well-calibrated confidence intervals that guarantee to include the true probabilities for each class output to occur\cite{Vovk:2005:ALR:1062391}. The confidence intervals for a test input are generated by considering the class distribution of labeled inputs assigned to the same category that are collected offline and are available to the system. In the literature, VP implementations use Support Vector Machines (SVMs) or DNN classifiers to create categories of labeled data\cite{Vovk:2005:ALR:1062391,lambrou2015inductive,papadopoulos2013reliable}. The additional problem we are considering is the computation of appropriate embedding representations that can lead to more efficient VPs. The main idea is to use distance metric learning and enable DNNs to learn a lower-dimensional representation for each input on an embedding space where the Euclidean distance between the input representations is a measure of similarity between the original inputs themselves. Using such representations we define taxonomies to form categories of similar input data. This not only reduces the memory requirements but is also more efficient in producing more informative intervals. 

%This idea is applied in the definition of taxonomies that create more accurate categories of similar data.
% If the computed probability intervals are too wide or lower than a desired prediction confidence, an alarm can be raised
% to indicate the need for human intervention.
% Real-time execution is important in many practical applications and the biggest execution time bottleneck of using VPs is the requirement of retraining an underlying model after every new input. In our work we overcome this challenge by using the Inductive Venn Prediction (IVP) framework which only require the training of the underlying model once, before the test execution. Evaluation of the method must be based on metrics that quantify the error rate and the quality of the probability intervals. For real-time operation, the time and memory requirements for computing the probability intervals must not add significant overhead on top of the existing DNN classifier.
% \input{siamese.tex}
\section{Probability Intervals based on Distance Metric Learning}
\label{sec:ivp}
Venn Predictors is a machine learning framework that can be combined with existing classifier architectures for producing well-calibrated multi-probability predictions under the IID assumption~\cite{balasubramanian2014conformal,Vovk:2005:ALR:1062391}. This means that the confidence assigned to a prediction is a probability distribution which in effect defines lower and upper bounds regarding the probability of correctness for all possible classes. VPs are well-calibrated and the probability bounds asymptotically contain the corresponding true conditional probabilities (proof in~\cite{Vovk:2005:ALR:1062391}). However the framework is computationally inefficient as it requires training the underlying algorithm after every new test input. Computational efficiency can be addressed using the Inductive Venn Predictors~\cite{lambrou2012reliable,lambrou2015inductive}, an extension of the VP framework.

Central to the VP and IVP frameworks is the definition of a Venn taxonomy. This is a way of clustering data points into a number of categories according to their similarity and is based on an underlying algorithm. For example a taxonomy can be defined to put in the same category examples that are classified in the same class by a DNN. The main idea of our approach is that the taxonomy can be defined efficiently by learning embedding representations of the inputs for which the Euclidean distance is a measure of similarity. To compute the embedding representations of the inputs we train a \textit{siamese network} using contrastive loss~\cite{hadsell2006dimensionality,koch2015siamese}.

We consider the training examples, $z_1,\dots,z_l$ from $\bm{Z}$, where each $z_i$ is a pair $(x_i,y_i)$ with $x_i$ the feature vector and $y_i$ the corresponding label. We also consider a test input $x_{l+1}$ which we wish to classify. IVP assumes that all the examples $z_1,\dots,z_{l+1}$ are independent and identically distributed (IID) generated from the same but usually unknown probability distribution. The available training examples are split into two parts: the \textit{proper training set} with $q$ examples and the calibration set with $l-q$ examples. The examples in the proper training set are used to train the siamese network which is used to define different Venn taxonomies. The roll of the taxonomy is to divide the $l-q$ calibration examples into a number of categories based on their similarity. This process takes place during the design time. 

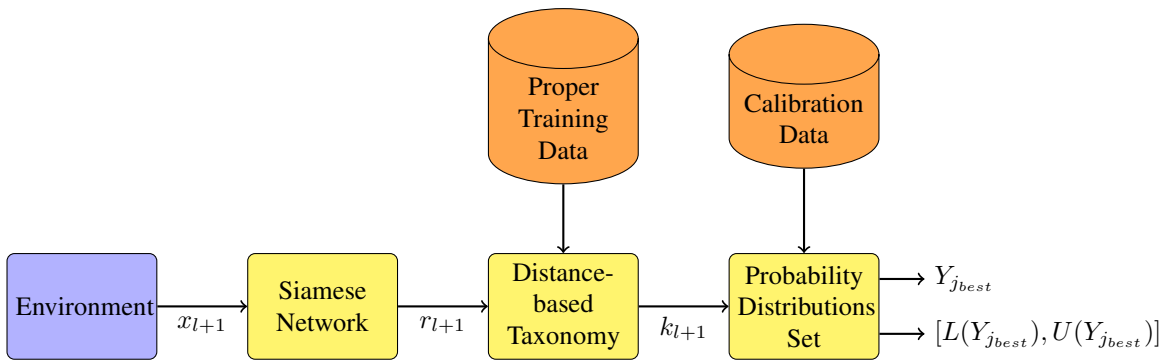
\begin{figure*}[ht]
\centering
% We need layers to draw the block diagram
% \pgfdeclarelayer{background}
% \pgfdeclarelayer{foreground}
% \pgfsetlayers{background,main,foreground}

% Define a few styles and constants
\tikzstyle{environment}=[draw, fill=blue!30, text width=5em,  
    text centered, minimum height=4em]
\tikzstyle{blocks}=[draw, fill=yellow!70, text width=5em, 
    text centered, minimum height=4em]
\tikzstyle{data_in}=[draw, cylinder, shape border rotate=90, fill=orange!70, text width=5em, 
    text centered, shape aspect=.4]

\begin{tikzpicture}
    \node (env) [environment, rounded corners=0.1cm] {Environment};
    \node [blocks, right of=env, node distance=3.2cm, align=center, rounded corners=0.1cm] (SiamNet) {Siamese\\Network};
    \node [blocks, right of=SiamNet, node distance=3.2cm, align=center, rounded corners=0.1cm] (knn) {Distance-based\\Taxonomy};
    \node [blocks, right of=knn, node distance=3.2cm, align=center, rounded corners=0.1cm] (distributions) {Probability\\Distributions\\Set};
    \node [data_in, above of=distributions, node distance=2.5cm, align=center] (cal_data) {Calibration\\Data};
    \node [data_in, above of=knn, node distance=2.5cm, align=center] (train_data) {Proper\\Training\\Data};
    
    \draw[->,thick] (env) -- (SiamNet) node[midway,below] {$x_{l+1}$};
    \draw[->,thick] (SiamNet) -- (knn) node[midway,below] {$r_{l+1}$};
    \draw[->,thick] (knn) -- (distributions) node[midway,below] {$k_{l+1}$};
    \draw[->,thick] (cal_data) -- (distributions);
    \draw[->,thick] (train_data) -- (knn);
    \draw[->,thick] (distributions.20) -- +(0.6cm,0) node[right, text width=5em] {$Y_{j_{best}}$};
    \draw[->,thick] (distributions.-20) -- +(0.6cm,0) node[right] {$[L(Y_{j_{best}}), U(Y_{j_{best}})]$};
    
    % \begin{pgfonlayer}{background}
    %     % Compute a few helper coordinates
    %     \path (SiamNet.west |- train_data.north)+(-0.5,0.2) node (a) {};
    %     \path (knn.south -| knn.east)+(+0.3,-0.3) node (b) {};
    %     \path[fill=red!20!green!30,rounded corners=0.2cm, draw=black!50, dashed]
    %         (a) rectangle (b);
    % \end{pgfonlayer}
\end{tikzpicture}
\caption{IVP classifier based on distance metric learning}
\label{fig:venn_taxonomy_knn}
\end{figure*}

After placing the calibration data into categories using the underlying algorithm for the taxonomy, during execution time we consider a test input $x_{l+1}$ and place it in a category $k_{l+1}$. The true class $y_{l+1}$ is unknown and IVP computes a lower and an upper probability $[L(Y_j),U(Y_j)]$ for every possible class $j=1,\dots,c$ based on the number of samples of each class in $k_{l+1}$. The predicted class for the classification is computed as:
\begin{equation}
\label{eq:classification}
j_{best}=\text{arg}\max_{j=1,\dots,c}\overline{p(Y_j)}
\end{equation}
where $\overline{p(Y_j)}$ is the mean of the probability interval assigned to $Y_j$. Along with the class $Y_{j_{best}}$ the IVP framework outputs the probability interval $[L(Y_{j_{best}}),U(Y_{j_{best}})]$. The steps taking place during execution are illustrated in Fig.~\ref{fig:venn_taxonomy_knn}.

% so all possible classes $Y_j$ are considered candidate labels one after the other. The empirical probability assigned to each candidate class will be:
% \begin{equation}
% \label{eq:emprirical_probability}
% p(Y_j)=\dfrac{|\{(x^*,y^*)\in k_{l+1}:y^*=Y_j\}|}{|k_{l+1}|}.
% \end{equation}
% $k_{l+1}$ will always be non-empty as it will contain at least the new example $x_{l+1}$. This creates a probability distribution for the label $y_{l+1}$ computed as the ratio of the data belonging to each class in a category. By assuming all possible classes $Y_j$ as the true class one after the other in $k_{l+1}$, a lower and an upper probability is computed for each class. These are the two bounds that define the probability intervals $[L(Y_j),U(Y_j)]$. The predicted class for the classification is computed as:
% \begin{equation}
% \label{eq:classification}
% j_{best}=\text{arg}\max_{j=1,\dots,c}\overline{p(Y_j)}
% \end{equation}
% where $\overline{p(Y_j)}$ is the mean of the probability interval assigned to $Y_j$. Along with the class $Y_{j_{best}}$ the IVP framework outputs the probability interval $[L(Y_{j_{best}}),U(Y_{j_{best}})]$. By placing the new example to each of the $n$ categories, one at a time, we compute a set of probability distributions that compose the multi-probability prediction of the IVP $P_{l+1}^{k_i}=\{p^{k_i}(Y_j):k_i\in\{k_1,\dots,k_n\}, Y_j\in\{Y_1,\dots,Y_c\}\}$. That way, the probability intervals assigned to each class for each category as well as the class classification that occur in each category can be computed offline.

\section{Distance-based Taxonomies}
\label{sec:taxonomies}

As proved in\cite{Vovk:2005:ALR:1062391} the probability intervals assigned to each classification by the VP are well-calibrated regardless of the choice of the Venn taxonomy and this holds in practice for IVP as well\cite{lambrou2015inductive}. However, the choice of the taxonomy affects the efficiency of the IVP. The probability intervals are desirable to be relatively narrow to minimize the uncertainty in the probability of correctness as well as create better separation between the probabilities of each class. We propose four different Venn taxonomies based on distance metric learning. The first two taxonomies are based on a $k$-Nearest Neighbors classifier. The naive approach, that we call \textit{k-NN} $V_1$, trains a $k$-NN classifier using the embedding representations of the proper training set. Then the calibration data, as well as each new test input, are placed to a category that is defined by the $k$-NN prediction using the computed embedding representations. 
That is, for a data point $x_{l+1}$ that needs to be placed into a category, its embedding representation is computed using the siamese network, $r_{l+1}=\text{Net}(x_{l+1})$ and its $k$ nearest training data are found. Depending on the class $\hat{y}_{l+1}$ that most neighbors belong to, the data point is assigned to the category
\begin{equation}
\label{eq:knnv1}
k_{l+1}=\hat{y}_{l+1}.
\end{equation}
This taxonomy leads to a number of categories that is equal to the number of classes in the dataset. An extension of the previous taxonomy, \textit{k-NN} $V_2$, is also based on a $k$-Nearest Neighbors classifier. However, we attempt to more accurately split the data into categories by taking into account how many of the $k$ nearest training data points are labeled different than the predicted class. For a data point $x_{l+1}$ with embedding representation $r_{l+1}$ that needs to be placed into a category we compute the $k$-nearest neighbors in the training set and store their labels in a multi-set $\Omega$. The category where $x_{l+1}$ is placed is computed as:
\begin{equation}
\label{eq:knnv2}
k_{l+1}=\hat{y}_{l+1}\times(k-\left\lfloor{\frac{k}{c}}\right\rfloor)+|i\in\Omega:i\neq \hat{y}_{l+1}|
\end{equation}
where $\hat{y}_{l+1}$ is the $k$-NN classification of $r_{l+1}$, $k$ is the number of nearest neighbors and $c$ is the number of different classes. This taxonomy aims at further improving the similarity of the data in each category leveraging the classifier's confidence. It is expected that the more similar labeled neighbor training data points, the higher the chance of the corresponding class being the correct one. That way each category of \textit{k-NN} $V_1$ is further split into $k-\left\lfloor{\frac{k}{c}}\right\rfloor$ new categories.

The ability of siamese networks to create clusters of similar data can be used to further reduce the Venn taxonomy computational requirements when there is a large amount of training data. Each class cluster $i$ corresponding to class $Y_i$, $i=1\dots,c$ can then be represented by its centroid $\mu_{i}=\frac{\sum_{j=1}^{n_i}r_j^i}{n_i}$, where $r_j^i$ is the embedding representation of the  $ j^{th} $ training example from class $Y_i$ and $n_i$ is the number of training examples labeled as $Y_i$. We propose another family of taxonomies based on the \textit{Nearest Centroids}. The \textit{NC} $V_1$ places the calibration data as well as each new test input to a category that is the same as the class assigned to their nearest centroid. The category where an example $x_{l+1}$ is placed is computed as:
\begin{equation}
\label{eq:ncv1}
k_{l+1}=\text{arg}\min_{j=1,\dots,c}d(r_{l+1},\mu_{j})
\end{equation}
where $d$ the Euclidean distance. This leads to a number of categories that is equal to the number of classes in the dataset. An extension of this taxonomy, the \textit{NC} $V_2$, attempts to form more accurate categories by taking into account the classification confidence. We expect data points of the same class to be more similar to each other when their embedding representations are placed at similar distances to their class centroid. That way each category of \textit{NC} $V_1$ is further split into two categories based on how close an example $x_{l+1}$ is to its nearest centroid:

\begin{equation}
\label{eq:ncv2}
k_{l+1}=2\times\text{arg}\min_{j=1,\dots,c}d(r_{l+1},\mu_{j})+h,
\end{equation}

\begin{equation*}
h=\begin{cases}
0, &\text{if } d(r_{l+1},\mu_\text{min})\leq\theta\\
1, &\text{otherwise}
\end{cases}
\end{equation*}
where $\mu_\text{min}=\text{arg}\min_{j=1,\dots,c}d(r_{l+1},\mu_{j})$ is the distance to the nearest centroid and $\theta$ a chosen distance threshold.

% In Fig.~\ref{fig:scatter} cases where two new test inputs end up in the same categories in taxonomies $k$-NN $V_1$ and NC $V_1$ but in different categories in taxonomies $k$-NN $V_2$ and NC $V_2$. $k$-NN $V_2$ and NC $V_2$ are expected to be formed capturing sample similarities more accurately with the trade-off of containing less calibration data.

% \begin{figure}[ht]
% \centering
% \input{images/scatter_taxonomies.tikz}
% \caption{Test samples a and b end up in the same category in taxonomies $k$-NN $V_1$ and NC $V_1$ but in different categories in taxonomies $k$-NN $V_2$ and NC $V_2$.}
% \label{fig:scatter}
% \end{figure}

% \begin{figure}[ht]
% \centering
% \includegraphics[width=1\linewidth]{images/taxonomy_example.pdf}
% \caption{Example of a Venn taxonomy based on distance metric learning and a $k$-Nearest Neighbors classifier}
% \label{fig:venn_taxonomy_example}
% \end{figure}
\section{Evaluation Metrics}
\label{sec:evaluation_metrics}
The performance of IVP based on the proposed taxonomies is evaluated regarding the accuracy, calibration and efficiency. The objective is for the computed probability intervals to contain the true probability of correctness for each prediction. The probability interval for a given input $x$ with predicted class $\hat{y}$ is $[L(\hat{y}),U(\hat{y})]$. Equivalently, the probability that $\hat{y}$ is not the correct classification will be in the complimentary interval $[1-U(\hat{y}),1-L(\hat{y})]$, called \textit{error probability interval}. The true probability of correctness for a single prediction is unknown so the correctness of the computed intervals is evaluated over a number of samples. To do this we use the following metrics:
\begin{itemize}
    \item cumulative errors
    \begin{equation}
        E_n=\sum_{i=1}^n err_i,
    \end{equation}
    \begin{equation*}
    err_i=\begin{cases}
    1, &\text{if } \text{classification }\hat{y}_i \text{ is incorrect}\\
    0, &\text{otherwise}
    \end{cases}
    \end{equation*}
    
    \item cumulative lower and upper error probabilities
    \begin{align}
        LEP_n=\sum_{i=1}^n [1-U(\hat{y})], && UEP_n=\sum_{i=1}^n [1-L(\hat{y})]
    \end{align}
\end{itemize}

% \begin{figure*}
% \centering
% \subfloat[][GTSRB - Baseline v2]{%
% \input{images/GTSRB_v2_errors.tikz}
% }
% \subfloat[][GTSRB - Baseline $k$-NN v2]{%
% \input{images/GTSRB_knn_v2_errors.tikz}
% }
% \\
% \subfloat[][Fruits360 - Baseline v2]{%
% \input{images/fruits360_v2_errors.tikz}
% }
% \subfloat[][Fruits360 - $k$-NN v2]{%
% \input{images/fruits360_knn_v2_erros.tikz}
% }
% \caption{Caption}
% \label{fig:cumulative_errors}
% \end{figure*}

To compare the IVP implementations based on our proposed taxonomies with the baseline taxonomies, scalar metrics are used that represent the performance regarding accuracy, calibration, and efficiency. Unlike the NN classifiers that produce a single softmax probability for each class, the IVP framework produces probability intervals. For the computation of the evaluation metrics the probability assigned to a class $Y_j$ will be $\overline{p(Y_j)}$ like in (\ref{eq:classification}). 
The accuracy of an IVP implementation is evaluated as the number of correct classifications over the number of attempted classifications and it is computed as
\begin{equation}
    accuracy=1-\dfrac{E_n}{n}.
\end{equation}
An efficient, or informative, IVP is one that makes predictions  with small diameter probability intervals and their median is as close to zero or one. The most popular quality metrics for probability assessments are the negative log-likelihood (NLL) and the Brier score (BS)~\cite{VERIFICATIONOFFORECASTSEXPRESSEDINTERMSOFPROBABILITY}. 
NLL is the simplest out of the two and only considers the probability assigned to the predicted class in (\ref{eq:classification}). It is computed as
\begin{equation}
    NLL=-\sum_{i=1}^n\sum_{j=1}^c t_i^j \log(o_i^j),
\end{equation}
where $o_i^j=\overline{p(Y_j)}$ of example $i$ and $t_i^j$ the one-hot representation of the ground truth classification label $y_i$ of example $i$, that is
\begin{equation*}
    t_i^j=\begin{cases}
    1, &\text{if } \text{classification }y_i=Y_j\\
    0, &\text{otherwise}
    \end{cases}
\end{equation*}
This metric is minimized by producing intervals that are narrow and have median probability close to one assigned to the correct class. Computational issues may occur as the log score explodes if we observe an event that the classifier considers impossible. BS is computed as
\begin{equation}
    BS=\dfrac{1}{n}\sum_{i=1}^n\sum_{j=1}^c (o_i^j-t_i^j)^2
\end{equation}
This is, in effect, the mean squared error of the predictions. Unlike NLL, BS considers the probabilities assigned to all possible classes and will penalize probability intervals assigned to incorrect classes that are not close to zero. There are different views in the literature regarding which scoring rule is more appropriate. \cite{benedetti2010scoring} emphasizes in the importance of the locality property, meaning, the scoring rule should only depend on the probability of events that actually occur and only NLL satisfies this. On the other hand, \cite{selten1998axiomatic} states that a scoring rule should be symmetric and only BS satisfies this. This means that if the true class probability is $p$ and the predicted probability is $\hat{p}$, then the score should be equal to the case where the true probability is $\hat{p}$ and the predicted probability is $p$. However, we think that both metrics produce useful insights in probability assessment so both are reported in our experiment results. 
The interval size has a significant role on how informative and interpretable a prediction is. We evaluate the size of the probability intervals by computing the average interval diameter as
\begin{equation}
D=\dfrac{\sum_{i=1}^n U(\hat{y})-\sum_{i=1}^n L(\hat{y})}{n} 
\label{eq:diameter}
\end{equation}

A well-calibrated IVP computes probability intervals that are representative of  the true correctness likelihood. Formally a model is well-calibrated when 
\begin{equation}
\mathbb{P}\left(\hat{y}=Y|\hat{p}=p\right)=p,\quad\quad\forall p\in[0,1]
\label{eq:calibration_definition}
\end{equation}
However, $\hat{p}$ is a continuous random variable so the probability in (\ref{eq:calibration_definition}) cannot be approximated using finitely many samples. According to (\ref{eq:calibration_definition}) a measure of miscalibration can be expressed as $\underset{\hat{p}}{\mathbb{E}}\left[\left|\mathbb{P}\left(\hat{y}=y|\hat{p}=p\right)-p\right|\right]$. The \textit{Expected Calibration Error} (ECE)~\cite{naeini2015obtaining} computes an approximation of this expected value across bins:

\begin{equation}
\text{ECE}=\sum_{m=1}^M \dfrac{|B_m|}{n}\left|\text{acc}(B_m)-\text{conf}(B_m)\right|
\label{eq:ECE}
\end{equation}
where $|B_m|$ is the number of samples in bin $B_m$, $n$ is the total number of samples and acc$(B_m)$ and conf$(B_m)$ are the accuracy and confidence of bin $B_m$ respectively as defined in~\cite{naeini2015obtaining}. Many times in safety critical applications it is more useful to compute the maximum miscalibration of a model than the mean value. This metric is called Maximum Calibration Error (MCE)~\cite{naeini2015obtaining} and is computed as:
\begin{equation}
\text{MCE}=\max_{m\in \{1,\dots,M\}} \left|\text{acc}(B_m)-\text{conf}(B_m)\right|
\label{eq:MCE}
\end{equation}
\section{Evaluation}
\label{sec:evaluation}
In this section, we evaluate the IVPs that use distance-based taxonomies with regard to accuracy, calibration, and efficiency. Additionally, for the evaluation of our proposed taxonomies, we use metrics regarding the performance of the siamese network in clustering similar input data, the execution time of the framework, and the required memory. 

\subsection{Experimental Setup}
The embedding representation computations, part of our proposed taxonomies, are not application-specific and can improve the performance of IVP in cases where inputs are high-dimensional. We evaluate the performance of IVP with distance-learning in two different classification problems. First, we have two case studies in image classification. The German Traffic Sign Recognition Benchmark (GTSRB) dataset is a collection of traffic sign images to be classified in 43 classes~\cite{GTSRB_cite}. The labeled sign images are of various sizes between 15x15 to 250x250 pixels depending on the observed distance. We convert all the images to a fixed shape of 96x96 pixels. The second dataset is the Fruits360~\cite{murecsan2017fruit}. This dataset contains images of 131 different kinds of fruits and vegetables. The input data are used in their original size, 100x100 pixels. 
% For both datasets, we use MobileNets as a classifier and underlying model for its low latency and memory requirements. In the case of GTSRB we used width multiplier $\alpha=0.5$ while for the Fruits360 $\alpha=1$. In both datasets, the representations generated from the original image inputs are 1-D vectors of size 128.

The second classification problem we consider is the detection of botnet attacks in IoT devices. As part of the evaluation in~\cite{meidan2018n}, authors made available data regarding network traffic while infecting different common IoT devices two families of botnets. Mirai and BASHLITE are two common IoT-based botnets and their harmful capabilities are presented in~\cite{7971869}. In the dataset there are data for the following ten attacks:
\begin{itemize}
    \item BASHLITE Attacks
    \begin{enumerate}
        \item Scan: Scanning the network for vulnerable devices
        \item Junk: Sending spam data
        \item UDP: UDP flooding
        \item TCP: TCP flooding
        \item COMBO: Sending spam data and opening a connection to a specified IP address and port
    \end{enumerate}
    \item Mirai Attacks
    \begin{enumerate}
        \item Scan: Scanning the network for vulnerable devices
        \item Ack: Ack flooding
        \item Syn: Syn flooding
        \item UDP: UDP flooding
        \item UDPplain: UDP flooding with fewer options, optimized for higher PPS
    \end{enumerate}
\end{itemize}
Including the benign network traffic we approach this as a classification problem with eleven classes. The available data are in the form of 115 statistical features extracted from the raw network traffic. The same 23 features, presented in~\cite{meidan2018n}, are extracted from five time windows of the most recent 100ms, 500ms, 1.5sec, 10sec, 1min. The features summarize the traffic in each of these time windows that has (1) the same source IP address, (2) the same source IP and MAC address, (3) been sent between the source and destination IP address, (4) been sent between the source and destination TCP/UDP sockets. These features are computed incrementally and in real-time. 
% Since the input data are arranged in 1-D as vectors of 115 values we use a fully connected DNN with two hidden layers, the first has 10 units, and the second which produces the embedding representations has 32 units.

The available data are used throughout the evaluation process the same way in every dataset. 10\% of the data are taken out to be used for testing and the rest is the training set. The training set is then split into the proper training set and the calibration set. The proper training set is randomly chosen as 80\% of the training set and is used to train the underlying models and for the computation of the categories. The calibration set is the remaining 20\% of the available training data is used only to form the categories during the design time. The reported evaluation results are computed on the separate test set. All the experiments run in a desktop computer equipped with and Intel(R) Core(TM) i9-9900K CPU and 32 GB RAM and a Geforce RTX 2080 GPU with 8 GB memory.

\subsection{Baseline}
To understand the effect of the distance metric learning in IVP we compare it with approaches that use DNN classifiers as underlying algorithms. A variety of Venn taxonomy definitions based on DNNs is proposed in~\cite{papadopoulos2013reliable}. $V_1$ assigns two examples to the same category if their maximum softmax outputs correspond to the same class. $V_2$, divides the examples in the categories defined by $V_1$ into two smaller categories based on the value of their maximum softmax output. Their chosen threshold for the maximum output to create the two smaller categories is $0.75$. $V_3$ divides the examples in the categories defined by $V_1$ into two smaller categories but this time based on the second highest softmax output. Their chosen threshold for the second-highest output is $0.25$. $V_4$ divides each category of taxonomy $V_1$ in two, based on the difference between the highest and second-highest softmax outputs. The threshold for this difference is $0.5$. In the same paper, they proposed a fifth taxonomy that creates the categories based on which classes have softmax outputs above a certain threshold. This taxonomy creates $2^C$ number of categories making its use infeasible in our evaluation datasets.

% The first baseline taxonomy is denoted as $V_1$. This is the simplest among the defined taxonomies and it assigns two examples to the same category if their maximum outputs correspond to the same class. This produces $c$ categories. The second baseline taxonomy, denoted as $V_2$, divides the examples in the categories defined by $V_1$ into two smaller categories based on the value of their maximum output. The used threshold for the maximum output that is used to create the two smaller categories will be $0.75$ and this leads to $2c$ classes. The third baseline taxonomy, denoted as $V_3$, divides the examples in the categories defined by $V_1$ into two smaller categories but this time this is based on the second highest output. The used threshold for the second highest output will be $0.25$ and this leads to $2c$ classes. The fourth baseline taxonomy, denoted as $V_4$, again further divides each category of taxonomy $V_1$ in two, based on the difference between the highest and second highest outputs. The threshold for this difference will be $0.5$ and this also leads to $2c$ classes. In the same paper they proposed a fifth taxonomy that creates the categories based on which classes have probabilities above a certain threshold. This taxonomy creates $2^C$ number of categories making its use infeasible in our evaluation datasets.

\subsection{Evaluation Results}
The difficulty to assign an input to a category and the memory demands increase as the size and complexity of the inputs increases. 
% A siamese network is used to map the input images into embedding representation vectors of size 128. 
Our goal is to evaluate our method using general-purpose and lightweight DNNs. For the image classification problems, we use the MobileNet architecture for both the embedding representation computation as well as the classifier used for the baseline taxonomies for its low latency and low memory requirements. The trade-off between accuracy and latency is configured by the hyperparameter $\alpha$. We set $\alpha=0.5$ in the case of GTSRB and $\alpha=1$ for the Fruit360. In both cases the embedding representation vectors are of size 128. In the case of the botnet attacks detection, the input data are arranged in vectors of 115 values so we use a fully connected DNN with two hidden layers, the first has 10 units, and the second which produces the embedding representations has 32 units.

\begin{figure}[ht]
\centering
\begin{tikzpicture}
\begin{axis}[
    % title=Inv. cum. normal,
    xlabel={sample \#},
    ylabel={Cumulative error},
    legend pos=north west,
    ymajorgrids=true,
    grid style=dashed
]
\addplot[name path=LEP, blue] table [x=sample, y=LEP, col sep=comma] {plot_data/GTSRB/test_nc_v2_ivp_errors_results.csv};
\addplot[name path=UEP, red] table [x=sample, y=UEP, col sep=comma] {plot_data/GTSRB/test_nc_v2_ivp_errors_results.csv};
\addplot[name path=CE, red, dashed, line width=1pt] table [x=sample, y=CE, col sep=comma] {plot_data/GTSRB/test_nc_v2_ivp_errors_results.csv};
\addlegendentry{CE};
\addplot[blue!70, opacity=0.1] fill between[of=LEP and UEP];
\legend{LEP,UEP,CE}
% \addplot[black] 
% 	coordinates {(2600,50) (2800,50)};
% \addplot[black] 
% 	coordinates {(2600,12.1) (2800,12.1)};
% \draw[<->,red] (axis cs:2800,12.1) -- (axis cs:2800,50);
% \node[right] at (axis cs:2800,25) {Uncertainty};
\end{axis}
\end{tikzpicture}
\caption{Illustration of cumulative metrics in the GTSRB dataset using the NC $V_2$ taxonomy}
\label{fig:cumulative_errors}
\end{figure}
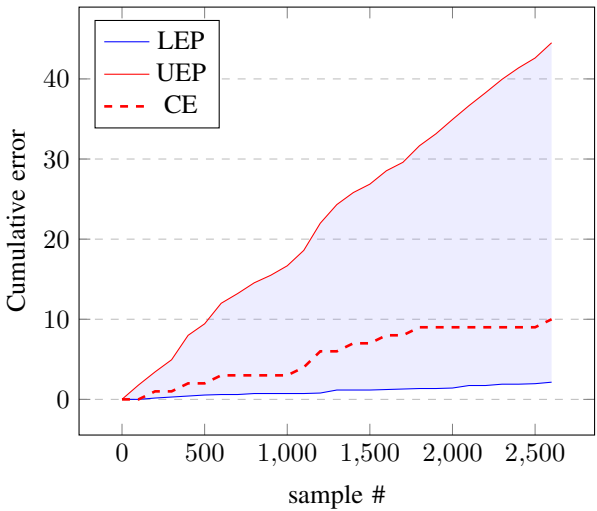

After training the siamese network and before it is used as part of the taxonomies we need to evaluate how well it performs in clustering similar inputs. For comparison, we use the embedding space produced by the penultimate layer of the DNN classifier~\cite{hinton2007learning}. A commonly used metric of the separation between class clusters is the \textit{silhouette coefficient}~\cite{ROUSSEEUW198753}. This metric evaluates how close together samples from the same class are, and far from samples of different classes and takes values in [-1,1]. The results on the silhouette analysis for the test inputs from both datasets are shown in Table~\ref{tab:silhouette}. The siamese network produces representations that are well clustered based on their similarity and better than the representations produced by the classifier DNN. This is important for constructing efficient categories using our proposed distance-based taxonomies.
% \vspace{-0.2cm}
\begin{table}[H]
\centering
\caption{Silhouette Coefficient Comparison}
\label{tab:silhouette}
\begin{tabular}{c|c|c|}
\cline{2-3}
                                & Classifier Embeddings & Siamese Embeddings \\ \hline
\multicolumn{1}{|c|}{GTSRB}     & 0.56                  & 0.98               \\ \hline
\multicolumn{1}{|c|}{Fruits360} & 0.52                  & 0.85               \\ \hline
\multicolumn{1}{|c|}{Ecobee Thermostat} & 0.27                  & 0.46               \\ \hline
\end{tabular}
\end{table}
% \vspace{-0.2cm}
For illustration, the cumulative upper and lower error probabilities as well as the cumulative error are plotted on the same axis in Fig.~\ref{fig:cumulative_errors} using the NC $V_2$ taxonomy and test data from the GTSRB dataset. The computed probability intervals successfully bound the true error-rate.

% For each sample, we first compute 
% the mean distance between $i$ and all other data points in the same cluster in the embedding space
% $$a(i)=\frac{1}{|C_i|-1}\sum_{j\in C_i, i\neq j}d(i,j) \text{  .}$$Then we compute the smallest mean distance from $i$ to all the data points in any other cluster
% $$b(i)=\min_{k\notin i}\frac{1}{|C_k|}\sum_{j\in C_k}d(i,j) \text{  .}$$
% The silhouette value is defined as
% $$s(i)=\frac{b(i)-a(i)}{\max\{a(i),b(i)\}} \text{  .}$$
% Each sample $i$ in the embedding space is assigned a silhouette value $-1\leq s(i)\leq 1$ depending on how close and how far it is to samples belonging to the same and different classes respectively. The closer $s(i)$ is to 1, the closer the sample is to samples of the same class and further from samples belonging to other classes. 

\begin{table*}[ht]
\centering
\caption{Evaluation metrics results}
\label{tab:taxonomy_comparison}
\begin{tabular}{|c|c|c|c|c|c|c|c|c|c|}
\cline{1-10}
Dataset  & Taxonomy & Accuracy & NLL & BS & D & ECE & MCE & Time & Memory \\ \hline\hline
\multicolumn{1}{|c|}{\multirow{8}{*}{GTSRB}} & $V_1$ & 0.994 & 111.835 & 0.013 & 3.29e-06 & 0.005 & 0.005& 3.6ms  &11.2MB\\ \cline{2-10}
\multicolumn{1}{|c|}{}  & $V_2$ & 0.992 & 58.104 & 0.055 & 5.30e-06 & 0.011 & 0.583& 6.6ms  &11.2MB\\ \cline{2-10}
\multicolumn{1}{|c|}{}  & $V_3$ & 0.993 & 75.394 & 0.038 & 4.48e-06 & 0.008 & 0.750& 3.7ms  &11.2MB\\ \cline{2-10}
\multicolumn{1}{|c|}{}  & $V_4$ & 0.991 & 70.279 & 0.053 & 5.06e-06 & 0.009 & 0.750& 2.9ms  &11.2MB\\ \cline{2-10}
\multicolumn{1}{|c|}{}  & $k$-nn $V_1$ & 0.998 & 41.575 & 0.005 & 3.30e-06 & 0.004 & 0.004& 3.2ms  &19MB\\ \cline{2-10}
\multicolumn{1}{|c|}{}  & $k$-nn $V_2$ & 0.998 & 41.126 & 0.005 & 3.30e-06 & 0.004 & 0.004& 3.6ms  &19.8MB\\ \cline{2-10}
\multicolumn{1}{|c|}{}  & NC $V_1$ & 0.998 & 41.575 & 0.005 & 3.30e-06 & 0.004 & 0.004& 2.9ms  &3.9MB\\ \cline{2-10}
\multicolumn{1}{|c|}{}  & NC $V_2$ & 0.996 & 38.444 & 0.046 & 6.21e-06 & 0.007 & 0.500& 3ms  &3.9MB\\ \hline\hline
\multicolumn{1}{|c|}{\multirow{8}{*}{Fruits360}} & $V_1$ & 0.983 & 1089.938 & 0.043 & 1.19e-06 & 0.008 & 0.113& 4.4ms  & 41MB\\ \cline{2-10}
\multicolumn{1}{|c|}{}  & $V_2$ & 0.986 & 816.893 & 0.144 & 1.57e-06 & 0.013 & 0.407& 4ms  &41.2MB\\ \cline{2-10}
\multicolumn{1}{|c|}{}  & $V_3$ & 0.985 & 870.470 & 0.154 & 1.55e-06 & 0.012 & 0.392& 4.6ms  &41.2MB\\ \cline{2-10}
\multicolumn{1}{|c|}{}  & $V_4$ & 0.985 & 836.295 & 0.159 & 1.58e-06 & 0.013 & 0.384& 2.7ms  &41.2MB\\ \cline{2-10}
\multicolumn{1}{|c|}{}  & $k$-nn $V_1$ & 0.993 & 532.314 & 0.025 & 1.19e-06 & 0.010 & 0.073& 3.3ms  &127.5MB\\ \cline{2-10}
\multicolumn{1}{|c|}{}  & $k$-nn $V_2$ & 0.993 & 466.311 & 0.088 & 1.42e-06 & 0.011 & 0.243& 3.7ms  &128.1MB\\ \cline{2-10}
\multicolumn{1}{|c|}{}  & NC $V_1$ & 0.991 & 605.087 & 0.027 & 1.19e-06 & 0.010 & 0.045& 3.6ms  &14MB\\ \cline{2-10}
\multicolumn{1}{|c|}{}  & NC $V_2$ & 0.988 & 725.556 & 0.208 & 2.22e-06 & 0.018 & 0.500& 3.5ms  &14.2MB\\ \hline\hline
\multicolumn{1}{|c|}{\multirow{8}{2cm}{\centering Ecobee Thermostat}} & $V_1$ & 0.823 & 4732.483 & 0.218 & 2.67e-08 & 0.003 & 0.009& 0.7ms & 52.2kB\\ \cline{2-10}
\multicolumn{1}{|c|}{}  & $V_2$ & 0.830 & 4310.008 & 0.200 & 4.20e-08 & 0.003 & 0.014& 0.7ms & 53.2kB\\ \cline{2-10}
\multicolumn{1}{|c|}{}  & $V_3$ & 0.830 & 4311.460 & 0.200 & 4.33e-08 & 0.002 & 0.015& 0.7ms & 53.2kB\\ \cline{2-10}
\multicolumn{1}{|c|}{}  & $V_4$ & 0.830 & 4306.791 & 0.200 & 4.26e-08 & 0.003 & 0.040& 0.6ms & 53.2kB\\ \cline{2-10}
\multicolumn{1}{|c|}{}  & $k$-nn $V_1$ & 0.935 & 2872.725 & 0.113 & 2.94e-08 & 0.001 & 0.003& 1.9ms & 43.8MB\\ \cline{2-10}
\multicolumn{1}{|c|}{}  & $k$-nn $V_2$ & 0.935 & 2299.023 & 0.096 & 1.33e-07 & 0.006 & 0.375& 2.4ms & 43.8MB\\ \cline{2-10}
\multicolumn{1}{|c|}{}  & NC $V_1$ & 0.794 & 5550.013 & 0.255 & 2.78e-08 & 0.006 & 0.017& 1ms & 24kB\\ \cline{2-10}
\multicolumn{1}{|c|}{}  & NC $V_2$ & 0.794 & 5541.171 & 0.255 & 4.02e-08 & 0.006 & 0.023& 0.9ms & 25kB\\ \hline
\end{tabular}
\end{table*}

The evaluation results are shown in Table~\ref{tab:taxonomy_comparison}. For both datasets, we observe that using the proposed distance-based taxonomies, IVP produces more accurate classifications. Even though the baseline $V_1$ taxonomy produces probability intervals that are as narrow as the intervals produced by some of the proposed taxonomies, the proposed taxonomies produce better quality intervals by keeping the intervals assigned to the correct class close to 1 and the intervals of the incorrect classes close to 0, as shown by the NLL and BS metrics. The differences in ECE are not significant but most of the proposed taxonomies produce probabilities that are better calibrated in the whole probability space $[0,1]$ with no areas of miscalibration as indicated by MCE.

The times required for the computation of a classification and the probability intervals when a new input arrives are similar in both the baseline and our proposed taxonomies and indicate they can be used for real-time operation. The speed bottleneck is the computations by the DNNs for either the classifications or the representation mapping. The $k$-NN computation step in the low-dimensional embedding representation space adds minimal overhead in the execution time. The memory requirements have two main parts: the memory required to store the DNN weights %by Keras - Tensorflow - you do not need this
and the memory required to store the categories after calibration.  The proposed taxonomies have the additional requirement to store either the embedding representations of the training data to be used by the $k$-NN or the centroid of each class. The representations of the training data are stored in a $k-d$ tree~\cite{Bentley:1975:MBS:361002.361007} for fast $k$-NN computation. With the use of low-dimensional representations, the additional memory required for the nearest centroid based taxonomies is small compared to the underlying DNN size.
% \vspace{-0.15cm}
\section{Conclusion}
\label{conclusion}
Although DNNs offer advanced capabilities, they must be complemented by engineering methods and practices for them to provide accurate measures of prediction confidence. For classification tasks, the IVP framework computes probability intervals that contain the probability of the prediction's correctness by examining the underlying model's accuracy on similar data. We presented computationally efficient algorithms based on appropriate embedding representations learned by siamese networks that make it possible for IVP to be used with high-dimensional data for real-time applications. The evaluation results demonstrate that the IVP framework using distance-based taxonomies produces high accuracy and probability intervals that are efficient and well-calibrated. Our choice of lightweight DNNs and small embedding representation size make the approach computationally efficient and can be used in real-time. A direction for future extension of this work is to improve the probability intervals, regarding their efficiency, during execution time. 
% how the computed probability intervals can be used to make decisions that satisfy predefined requirements.
% \vspace{-0.15cm}
\section*{Acknowledgment}
The material presented in this paper is based upon work supported by the Defense Advanced Research Projects Agency (DARPA) through contract number FA8750-18-C-0089. The views and conclusions contained herein are those of the authors and should not be interpreted as necessarily representing the official policies or endorsements, either expressed or implied, of DARPA.

\bibliographystyle{abbrv}
\bibliography{main}

\end{document}